# Intelligent 3D Network Protocol for Multimedia Data Classification using Deep Learning

Arslan Syed[1,†], Eman A. Aldhahri[2], Muhammad Munawar Iqbal[1], Abid Ali[1,3], Ammar Muthanna[4,5], Harun Jamil[6,†] and Faisal Jamil[7]

1. Department of Computer Science, University of Engineering and Technology, Taxila 47080, Pakistan; syedarsalan.uet@gmail.com (A.S.); munwariq@gmail.com (M.M.I.); abidali.hzr@gmail.com (A.A.)
2. College of Computer Sciences and Engineering, Department of Computer Science and Artificial Intelligence University of Jeddah, Saudi Arabia
3. College of Computer Sciences and Engineering, Department of Computer Science and Artificial Intelligence University of Jeddah, Saudi Arabia
4. Department of Computer Science, Peoples' Friendship University of Russia (RUDN University), 6 Miklukho-Maklaya St., 117198 Moscow, Russia; (A.C.)
5. Department of Telecommunication Networks and Data Transmission, The Bonch-Bruevich Saint-Petersburg State University of Telecommunications, 193232 Saint Petersburg, Russia
6. Department of Computer Engineering, Jeju National University, Jejusi 63243, Jeju Special Self-Governing Province, Republic of Korea
7. Department of ICT and Natural Sciences, Faculty of Information Technology and Electrical Engineering, Norwegian University of Science and Technology(NTNU), Larsgårdsvegen 2, Ålesund, 6009, Norway
\* Correspondence: faisal.jamil@ntnu.no
† These authors contributed equally to this work.

**Abstract:** In videos, the human's actions are of three-dimensional (3D) signals. These videos investigate the spatiotemporal knowledge of human behavior. The promising ability is investigated using 3D convolution neural networks (CNNs). The 3D CNNs have not yet achieved high output for their well-established two-dimensional (2D) equivalents in still photographs. Board 3D Convolutional Memory and Spatiotemporal fusion face training difficulty preventing 3D CNN from accomplishing remarkable evaluation. In this paper, we implement Hybrid Deep Learning Architecture that combines STIP and 3D CNN features to enhance the performance of 3D videos effectively. After implementation, the more detailed and deeper charting for training in each circle of space-time fusion. The training model further enhances the results after handling complicated evaluations of models. The video classification model is used in this implemented model. Intelligent 3D Network Protocol for Multimedia Data Classification using Deep Learning is introduced to further understand space-time association in human endeavors. In the implementation of the result, the well-known dataset, i.e., UCF101 to, evaluates the performance of the proposed hybrid technique. The results beat the proposed hybrid technique that substantially beats the initial 3D CNNs. The results are compared with state-of-the-art frameworks from literature for action recognition on UCF101 with an accuracy of 95%.

## 1. Introduction

In data mining and machine learning, the class imbalance problem has been addressed significantly by researchers. Whenever a dataset is not properly structured, its data instances are not well known and are significantly improved in real-world datasets. Majority classes and minority classes are defined within 3D datasets for multiple data. The identification of human behavior is growing in several applications, including video monitoring, connectivity between people and computer systems, and video recovery. In several current video applications, automated detection of events of interest is a crucial component. Therefore, creating modern behavior detection algorithms with greater precision and better handling diverse situations is often essential. Recognition of human behavior is a fundamental and daunting process that has been studied for decades.

It is motivated by the phenomenal performance of neural networks (CNNs), which detect visual representations in pictures. Several recent studies use deep models to create end-to-end networks to recognize human behavior in videos. Modeling Spatio-temporal knowledge together via a 3D CNN in a broad end-to-end network offers a simple and successful path to identifying behavior, given the performance increase that 3D CNN deep networks have made [1–3] in cases of volumetric analysis. The efficiency of video activity recognition in contrast with that obtained with 2D CNNs for visual recognition in images is far from satisfactory. A dual-stream architecture that leverages 2D CNNs pre-trained on big picture datasets offers the most state-of-the-art findings for action recognition [4,5]. However, this method does not consistently explain the preference of design [6]. Although, in the light of the latest 3D CNN action recognition networks, we remember the same design as in C3D [7,8], several of these approaches have the same 3D concentration layer by layer.

With the spatial and temporal signals being connected across each 3D convolution, the network of hundreds of 3D convolution layers becomes even more complicated to configure due to the exponential growth of the space for the 2D CNNs solution. The memory costs for 3D convolution are too high for constructing a 3D CNN that does not usually require the profound features of current 3D CNNs. For instance, a 3D CNN 11-layer needs almost 1.5 times more memory than a residual 152-layer simple network. Based on findings from the results, it is observed that reducing the number of 3D convolution layers saves memory, and growing the depth of the characteristic maps is an advantage for 3D CNNs. In our latest deep learning platform, we are proposing this research work with modern deep architecture to solve this issue and enhance the efficiency of 3D CNNs for action acknowledgment. In addition, this design makes it possible for the network to obtain improved efficiency with fewer spatial and temporal fusions at each spatial and time stage. It decreases the difficulty of each spatial and temporal fusion round by utilizing the residual cross-domain relation.

Multimedia data are commonly defined as text, images, audio, and video. The surge of sensor-enriched mobile devices has turned video into a modern contact medium between users. Video data are produced, published, and distributed massively. It increases the requirements of bandwidth and storage capacity. It is an integral part of the big data of today. It has facilitated the creation of innovative technologies across a broad spectrum of applications. Video processing is used in a web advertisement, image analysis, and video tracking. This progress raises the question about these technological advancements in video production and comprehension. A significant challenge to be tackled for improved applications and standards is the definition of the extensive video toolkit. Within multimedia classification, the classifier for multimedia data is important in research. Classifying technological innovations and providing effective change in classification work is important.

Our work in this thesis represents one of the few studies showing very strong results. Our contributions are as follows:

- A fully convolutional 3D CNN pipeline for action/object segmentation in video and video classification. The approach leverages a model pre-trained on a large-scale action recognition task as an encoder to enable us to perform unsupervised video classification.
- We have contributed to using separable filters to reduce the standard 3D convolutions' computational burden significantly. It will also reduce the training and testing time.
- Our proposed deep video classification model with competitive performance incorporates spatial and temporal features with features extracted with the aforementioned 3D ConvNet. Extensive experiments on several benchmark datasets demonstrate the strength of our approach for Spatio-temporal action segmentation as well as video classification.

The rest of the paper is organized as follows. Section 2 describes the related work about feature extraction and 2D and 3D. Section 3 illustrates the materials and methods used in this study. Sections 4 and 5 demonstrate experimentation, implementation, results, and discussion with limitations of research. Finally, the whole research is concluded in Section 6.

## 2. Related Work

### 2.1. Deep learning and Convolutional Neural Network Architectures

Profound learning is a big machine learning unit. It is centered on a sequence of algorithms to model a high degree of data abstraction, as shown in Figure 1. Deep learning is a recently established area that is commonly employed in the area of pattern analysis and self-learning. It consists of several layers of processing composed of linear and non-linear transformations [9]. Deep learning converts data into layers in contrast to most simplistic learning algorithms. A convolutional neural network is a variant of a neural network that contains convolutional layers, pooling layers, fully connected layers, and normalization layers. This distinct arrangement of layers makes convolutional neural networks exceedingly better at image recognition tasks than other networks.

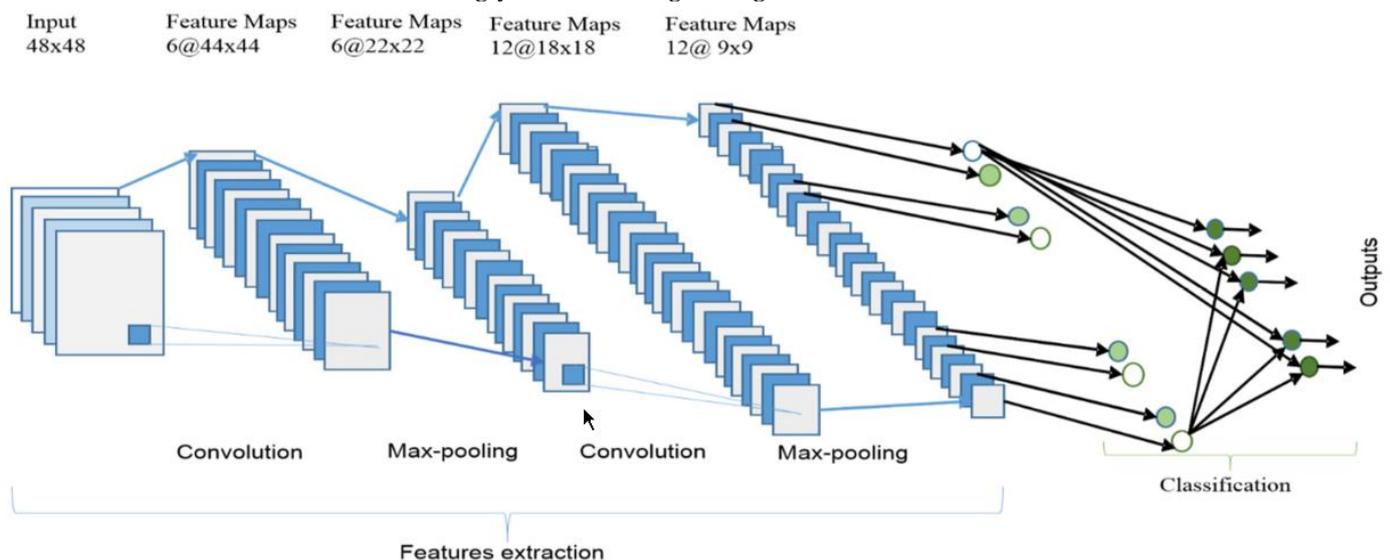

**Figure 1.** Convolutional Neural Network Architecture [10].

The human and animal interactive cortex are today's most robust and dynamic forms of vision. The visual cortex is formed by the proliferation of various cells that are very sensitive to specific areas of the eyes and is recognized as the receptive field of the trim zone. It is designed to occupy the whole visual area in a sensitive environment. Two distinct cell groups are found in this visual cortex. This visual cortex structure encourages CNN.

2.1.1. VGG Net

Visual Geometry Group (VGG) [11 Net is a CNN architecture submitted at ImageNet Challenge in 2014 that secured the first position in localization and second in classification problems. VGG Net consists of 16–19 deep convolution layers with 3 × 3 filters. With this CNN model, we can achieve state-of-the-art results with any dataset.

VGG Net is publicly available in two versions, VGG16 and VGG19, for usage in computer vision.

2.1.2. Inception V3

Inception V3 [12] is one of the most efficient CNN architectures used for image classification is deep learning. Inception V3 was benchmarked on the ILSVRC 2012 challenge of image classification validation set and trained on the ImageNet dataset, which yielded much better results than VGG Net and Google Net. It was proposed in 2015 and utilized three Inception blocks within the convolution layers. This model is publicly available as an application and can be used for image classification purposes.

2.1.3. Xception Net

Xception et al. [13] is a CNN architecture inspired by Inception Architecture and uses depthwise separable convolutional layers followed by pointwise convolutional layers. It is just like Inception V3 [12], but depthwise detachable convolution units replace the Inception units. Xception et al. [13] tested on the ImageNet dataset, and it was slightly better in performance than Inception V3. Still, it was much better in performance and accuracy when trained on a larger image dataset. The dataset consisted of 350 million images of 17,000 classes. The number of parameters in Xception is the same as in Inception V3 but is better used to improve the performance.

2.1.4. Wavelet CNN

Wavelet CNN [14] is an architecture specially designed for texture classification. Like other CNN architectures, it treats the convolution and pooling layers as spectral analysis layers, which became very useful in texture classification in which other CNN architectures were failing. Wavelet CNN has utilized spectral analysis integrated into CNNs and analyzes the image in the frequency domain instead of the spatial domain as other CNNs do. As a result, Wavelet CNNs perform much better when tested against state-of-the-art CNN architectures such as VGG Net, AlexNet, and T-CNN in texture classification and use much less memory than these architectures.

2.1.5. Mobile Nets

Mobile Nets are CNN architectures specially designed for mobile and embedded systems using mobile vision software. These CNNs use depthwise separable streamlined architectures to become lightweight networks. These CNNs were trained on the ImageNet dataset and were found very accurate and efficient. In addition, these Nets allow the model builder to select between two hyperparameters for the appropriate network size for their application.

2.1.6. Dense Net

A dense Net or dense convolution network is a deeper network in which each layer is connected to every other layer, like a feed-forward network [15]. There are L(L + 1)/2 connections per layer. This architecture is specially designed to lower the number of parameters, have high efficiency, reuse feature maps, reduce vanishing gradient problems, and strengthen feature propagation. This model was trained on four commonly used datasets: ImageNet, Cifar-10, Cifar-100, and SVHN. Dense Net was very efficient and accurate on these datasets and required less memory and computations to achieve high performance. Dense Net is publicly available for usage in computer vision.

2.1.7. Deep Q-Networks

A Deep Q-Network (DQN) is a profound learning model that combines a deep CNN with Q-learning [16]. DQNs can learn directly from sensory stimuli, unlike earlier reinforcement learning agents. Preliminary findings were published in 2014, accompanied in February 2015 by a corresponding article. The thesis identified an Atari 2600 Game program. Other deep enhancement versions followed it [17].

2.1.8. Deep Belief Networks

The CDBNs have a somewhat comparable framework to CDBN networks and are equipped similarly to deep faith networks. The 2D picture form was then used, as with CNNs, and pre-training was used as deep belief networks. A basic framework was given, which can be used for several images and signal processing tasks using standard picture databases such as CIFAR test performance. Using CDBNs were collected [18].

2.1.9. Time-Delay Neural Networks

The first convolution network was the time-delay neural network (TDNN). TDNNs are networks of fixed-size convolution that exchange weights over time [19]. Time-invariantly, the speech signals may be interpreted similarly to the invariance of CNN's localization. At the beginning of the 1980s, they were launched. As a result, tiling neuron outputs may reach timed levels.

*2.2. Components and Tools*

There are different types of components developed by different vendors and research. Tensor Flow libraries are a powerful tool to execute deep neural networks. The optimizers are also used to enhance the performance of the deep network. Some of them are given as follows:

2.2.1. Tensor Flow

Tensor Flow is a large-scale deep machine learning library and system which runs on many operating systems such as Windows, Mac OSX, and Linux. Tensor Flow can be used to create neural network architecture such as CNN or RNN. It has a graphical system that maps the nodes in many devices such as CPUs, GPUs, and TPUs. In addition, TensorFlow supports a variety of training algorithms and applications, which aids in developing various architectures for deep learning and their optimizations. TensorFlow is an open-source project and is used by many Google services. Due to its performance and usability, it can be used in several applications.

2.2.2. Long Short-Term Memory Units (LSTMs)

In the mid-90s German researchers Sepp Hochreiter and Juergen Schmid Huber suggested a variant of the repeating system with so-called large short-term memory units (LSTMs) to address the issue of the fading gradient [14]. The support of LSTMs conserves errors that can be recorded over time and layers. They encourage repeat networks to keep learning more than 1000-time steps by retaining a more constant error and thus open a channel to connect causes and results remotely. It is one of the main challenges for machine learning and AI, as algorithms always face conditions where incentive signals such as existence are scarce and slow. This same dilemma has been discussed by philosophers, who have theorized unseen and abstract effects of karma or God's compensation for our acts.

2.2.3. ADAM Optimizer

ADAM is an algorithm for stochastic gradient optimization. ADAM is very efficient, requires less computation, is easy to implement, requires less memory, and is very suitable for extensive data or data with many parameters [16]. This algorithm also suits problems having sparse or noisy gradients. This method has satisfactory results compared to other stochastic gradient methods.

*2.3. Usefulness of RNNs*

A video is just an ordered sequence of frames. Frames contain spatial information, and the sequence of those frames contains temporal information. Convolutions may be used for spatial processing and the recurrent layers in an RNN (recurrent neural network) for temporal processing. RNNs utilize internal state (memory) to process variable-length inputs. Video data are of variable length, usually because the number of frames in each video differs. So, it is natural to use an RNN for such a task at hand, but it is challenging to train an RNN. Another critical term to mention here is LSTM (Long Short-Term Memory). An LSTM is an RNN with feedback connections and can process entire data sequences (video). Classifying, making predictions, and processing a time series data of unknown duration is a well-suited task for an LSTM. Because LSTMs are not hardware friendly, they are being replaced by attention-based networks in most of the recent implementations by major companies such as Google, Facebook, and Salesforce, to name a few.

*2.4. 3D Convolutions for Video Data*

For video classification, 3D ConvNets is an inspiring option because they implement convolutions (and max pools) implicitly in 3D space, where the third dimension is time in our case. (Technically, it is 4D as our 2D artifacts are shown as 3D vectors, but the net effect is identical). (a) 2D image convolution results in a frame. (b) The application of 2D convolution (multiple images as multiple channels) to a video volume often results in a shot. (c) The operation of 3D convolution on a video volume contributes to a new volume retaining the time input details seen in Figure 2. 3D CNNs are better suited to learn Spatio-temporal features than a 2D Convent because 3D CNNs can retain the temporal information owing to 3D convolutions. 3D CNN pooling and convolutions are carried out spatiotemporally, while a 2D Convent only outputs an image when used on an image (Figure 2 a,b). 3D convolutions preserve temporal information (Figure 2c).

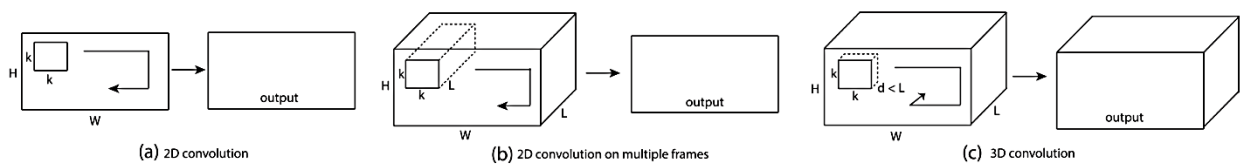

**Figure 2.** 2D vs. 3D convolution (**a**) 2D convolution (**b**) 2D Convolution on Multiple Frames and (**c**) 3D convolution.

3D CNNs are being used to create models that enhance the detection of smartphones and hand-held guns from video data such as surveillance footage. In addition, such 3D CNNs are also being used to help medical scans detect and segment tumors in medical footage, such as MRI scans. Due to their sophistication, 3D CNNs being used now are at an early level, but the advantages they may offer are worth studying.

The traditional method for video classification [13,14,17,20] includes three key phases: First, thinly [21] or in a limited collection of points of interest [11,22], local visuals that represent an area of the video are produced. Then, the characteristics are merged in a fixed-scale video overview. One standard method is quantifying and aggregating visual terms in histograms in various spatial-temporal locations and areas in the video utilizing a studied k-means-dictionary [23].

A classifier such as SVM is taught to differentiate between the visual groups of interest in the resulting "sack of terms" description. Convolution neural networks [12] are a biologically based profound learning model that substitutes all three steps with a neural network equipped from raw to classifying output values end to end. Regulatory regularization by restricted interconnections of layers (local filters), exchanging parameters (convolutions), and unique local neurons (total grouping) are specifically used for the spatial structure of photos [24]. Thus, the necessary architecture moves from role design and aggregation techniques to the network communication framework and hyperparameter choices [25]. Until recently, CNNs were implemented on relatively limited scales (MNIST, CIFAR10/100, NORB, Caltech-101/256) due to computational constraints. Still, improvements to GPU hardware allowed CNNs to scale networks of millions of parameters that have, in turn, led to substantially improved imaging classifications [1] and object-sensing [16,26].

Moreover, the features gained via broad ImageNet-qualified networks have shown that many regular image recognition datasets deliver state-of-the-art output when labeled with an SVM, even without a fine tone [27]. From a size perspective and dimensionality, video data are fascinating. Instead of a single image being classified with the help of a CNN, we have a large stack of images with temporal relations intertwined. Using a CNN for such data will not yield desired results, so a different approach such as a 3D CNN would be better and suffice. Although there is a scarcity of a wide range of courses shared with all popular implementations of CNNs in imaging fields, we believe that this is partly attributed to the absence of broad video classification benchmarks [8]. Used databases in particular (KTH, Weizmann, UCF Sports, IXMAS, Hollywood 2, UCF-50) include just up to several thousand clips and up to a few decades in grades. Several instances and their variants have the most significant current datasets like CCV (9317 videos and 20 classes). The newly released UCF-101 (13,320 videos and 101 classes) also has a dwarf of picture datasets [18]. Given these limits, many CNN [28] extensions have been investigated in the video domain. Paper [29] and [6] expand the CNN picture to visual realms by considering space and time as the corresponding input dimensions and conducting time and space convolutions. These extensions are known to be just one logical generalization. Spatiotemporally training systems based on convolutional gated restricted Boltzmann [19] and independent subspace analysis [30] have also been created.

*2.5. Feature Extraction and Classification Methods*

The computing culture in vision has been researching videos for decades. Different challenges have been addressed over the years, including intervention recognition [20], incident identification [31], video recovery [29], tracking of incidents and activities, and many more. Video depictions are a large part of these productions. Laptev et al. [20] suggested the application of Harris corner detectors to 3D Spatio-temporal points of concern (STIPs). For action detection, SIFT and HOG are generalized to SIFT-3D and HOG3D [15] as well. Dollar et al. ActionBank has suggested Sadanand and Corso have founded cuboid characteristics for conduct identification for action recognition. Wang et al. [32] recently suggested an updated, state-of-the-art handcrafted Dense Trajectory (iDT).

The iDT descriptor indicates that temporal signals can be treated differently from spatial signals. It begins with densely sampled practical points in video frames instead of expanding the Harris corner detector into 3D and uses optical flows to track these. Various customized features are removed from the road for each tracker corner. This process is computer-intensive and can be impeded by massive datasets, despite its good efficiency. Recently, efficient parallel machines (GPUs, CPU clusters) and vast quantities of training data have rendered the comeback of transforming neural networks (ConvNets) a success in visual recognition [6]. ConvNets are also discussing the issue of human pose prediction in photographs and videos. These deep networks are more critical for learning picture features [5]. Zhou et al. were similarly good at the moved learning activities. For visual feature research, profound learning has been used in unattended settings [33]. The authors used ISA stacked to analyze Spatio-temporal functions for images [34]. While this approach has produced strong results in identifying behavior, it is still complex in quantitative exercises and challenging to scale with big datasets. For human behavior [12] and medical imaging [30], 3D ConvNets were suggested. 3D convolution was often used for studying Spatio-temporal features with Limited Boltzmann machinery.

Karpathy et al. [15] recently equipped deep video recognition networks with a broad video dataset. Simonyan et al. [35] used two-stream networks to obtain the maximum performance in detecting behavior. W. Jiang et al. [12] proposed the method used by 3D ConvNets is more strongly associated with us. This system used a camera tracker for the human subjects and head monitoring for the section. For behavior description, the segmented video volumes are used as inputs for a three-layer ConvNet. Our system, in comparison, uses complete video frames as inputs, relies on no preprocessing, and is simple to scale to broad datasets. We also have some parallels with [27,35] in using complete ConvNet files.

Nevertheless, these approaches have been developed only through 2D conversion and 2D pooling (except for Sluggish Fusion in [27]), while our model carries out 3D convolutions and 3D bundling that spreads time over all the network's layers (see Section 3 for further details). We also prove that pooling space, time, and deeper networks increasingly yields improved outcomes. We assume 3D ConvNet is suitable for space-time computing.

The results are compared to 2D ConvNet. Due to 3D convolution and 3D pooling, 3D ConvNet can model temporal details more easily. 3D ConvNets conduct Spatio-temporal convolution and pooling in 2D ConvNets only in space. Table 1 demonstrates the discrepancy in two-dimensional representations applied to images that can generate a picture. Therefore, 2D ConvNets lose temporal details on an input signal after all the convolution operations. The time information of the input signals that generate an output volume is retained only in 3D convolution. For 2D and 3D pooling, the same phenomenon exists.

In [35], while several frames are used as input for a temporal stream network, temporal knowledge drops totally after the first convolution layer due to the 2D convolutions. Infusion models have used 2D convolutions, and after their first convolution layer, most networks lose their temporal data. The Sluggish Fusion model in [27] uses average pooling and 3D convolutions in the first three layers. It is the primary explanation of why all the networks analyzed were the best [27]. After the third convolution sheet, however, it still loses all-time details. We are attempting empirically to define a successful 3D ConvNets architecture. Although training deep networks on large-scale video datasets is very time-consuming, we first try to find the exemplary architecture using UCF101, a medium-sized dataset. We review the results on a broad dataset for a limited number of network trials. Small reception areas of three to three convolution nodes with deeper architectures show better performance, according to the findings in 2D ConvNet. Therefore, we set the receptive spatial region for our architecture analysis to three to three and only vary the temporal depth of 3D kernels.

In [12], the 3D ConvNets method is more strongly associated with us. This system used a camera tracker for human subjects and head monitoring for the section. For the description of behavior, segmented video volumes are used as inputs for a three-layer ConvNet. Our system, in comparison, uses complete video frames as inputs, relies on no preprocessing, and is simple to scale to broad datasets.

**Table 1.** Critical analysis of state-of-art Video Analysis and Classification methods.

| References | Dataset | Data Augmentation | Model Selection | Traditional ML Classifier | Deep Learning Framework | Output Classes | Classification Accuracy |
|---|---|---|---|---|---|---|---|
| Donahue et al. [36] | UCF 101 | No | Yes | SVM | No | 101 | 82.9% |
| Srivastava et al. [35] | UCF 101 | No | Yes | Genetic Fuzzy | No | 101 | 84.3% |
| Wang et al. [32] | UCF 101 | No | No | SVM | - | 101 | 85.9% |
| Tran et al. [37] | UCF 101 | No | Yes | Decision Tree | - | 101 | 86.7% |
| Simonyan et al. [38] | UCF 101 | No | Yes | Fuzzy | - | 101 | 88.0% |
| Lan et al. [19] | UCF 101 | No | Yes | Random Forest | - | 101 | 89.1% |
| Zha et al. [39] | UCF 101 | No | No | SVM | - | 101 | 89.6% |

We also have some parallels with [27,35] in using complete ConvNet files. However, these approaches are based on only 2D and 2D (except in the Slow Fusion process) convolutions, whereas our approach offers a 3D and 3D conversion, propagating time details through all network layers. We also demonstrate that the incremental pooling of space, time, and deeper networks yields the best results, as shown in Table 1.

### 3. Materials and Methods

The first thing we do is to acquire one of the most challenging datasets of videos for action recognition (UCF 101). Figure 3 shows the proposed framework of the 3D CNN model. Learning from a large set of videos was troublesome for the available computer. Memory and computational power constrain the number of videos given to the 3D CNN for training and feature extraction. So, we split the dataset into batches for processing. Figure 4 shows Dataset Sample for UCF 101 Classes Training and Testing Split, where each batch contains five videos, as shown in Figure 5 and Section 3.2.

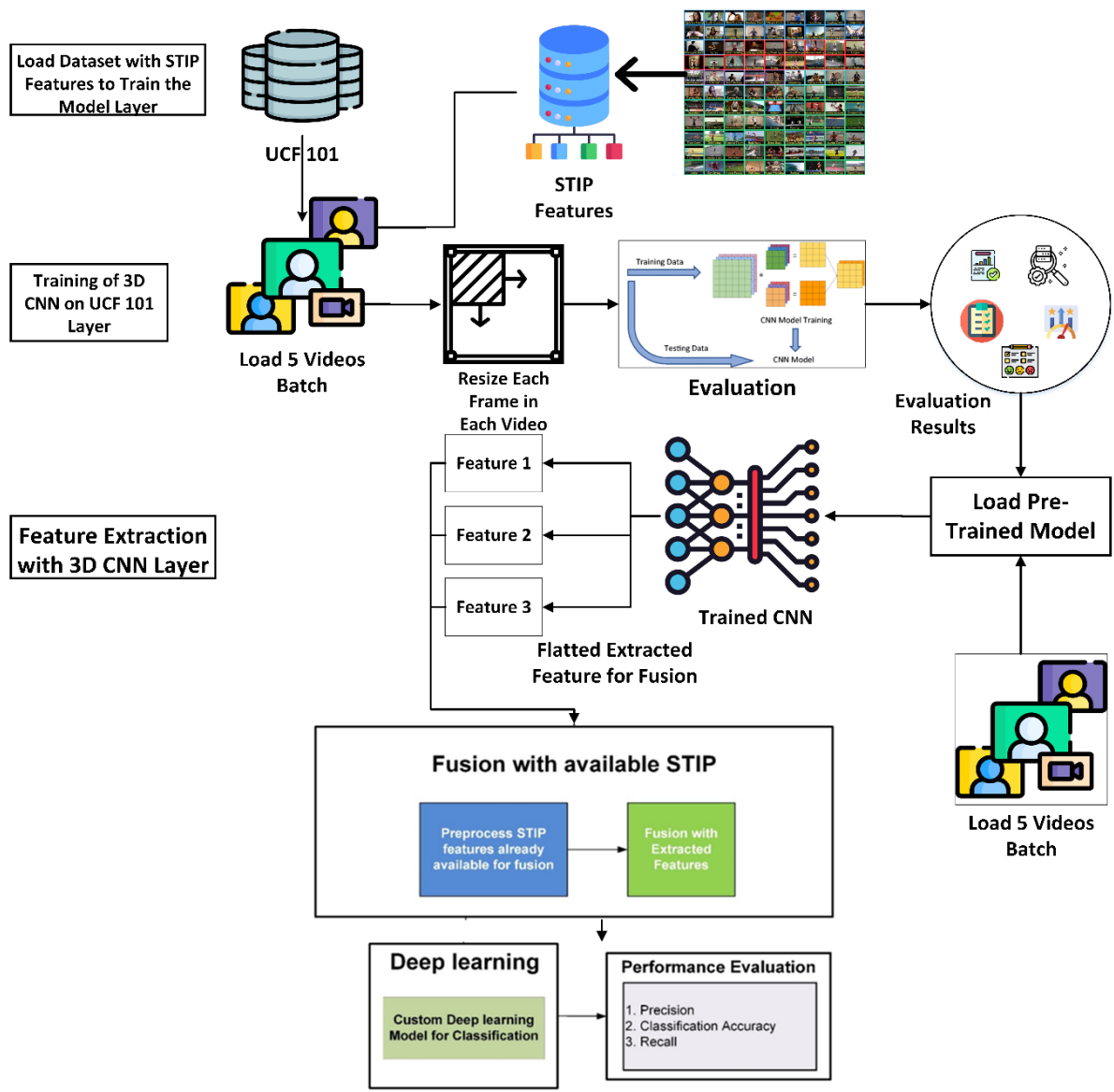

**Figure 3.** Proposed Research Methodology for 3D CNN.

*3.1. Dataset*

UCF101 is a collection of measurement identifications of practical YouTube action videos featuring 101 action groups. This data collection is an expansion of the 50 action groups in the UCF50 dataset. Figure 4 elaborates the UCF dataset. The UCF101 offers the most varied activities with 13,320 videos of 101 activity types and includes variations of camera movement, location, place, size of the target, angle, puzzled context, and lighting conditions. Although most existing action data identification sets are impractical and actor-based, UCF101 is intended to facilitate research into action recognition by learning and to discover different types of practical action. The videos are divided into 25 classes in 101 action categories, with 4–7 action videos per category probable. Videos of the same party may share some common characteristics, such as similar backgrounds or a similar outlook. Activity classification may be categorized into five types:

(1) Contact between people and things;
(2) Body-motion only;
(3) Contact between people and persons;
(4) Performing musical instruments;
(5) Sports.

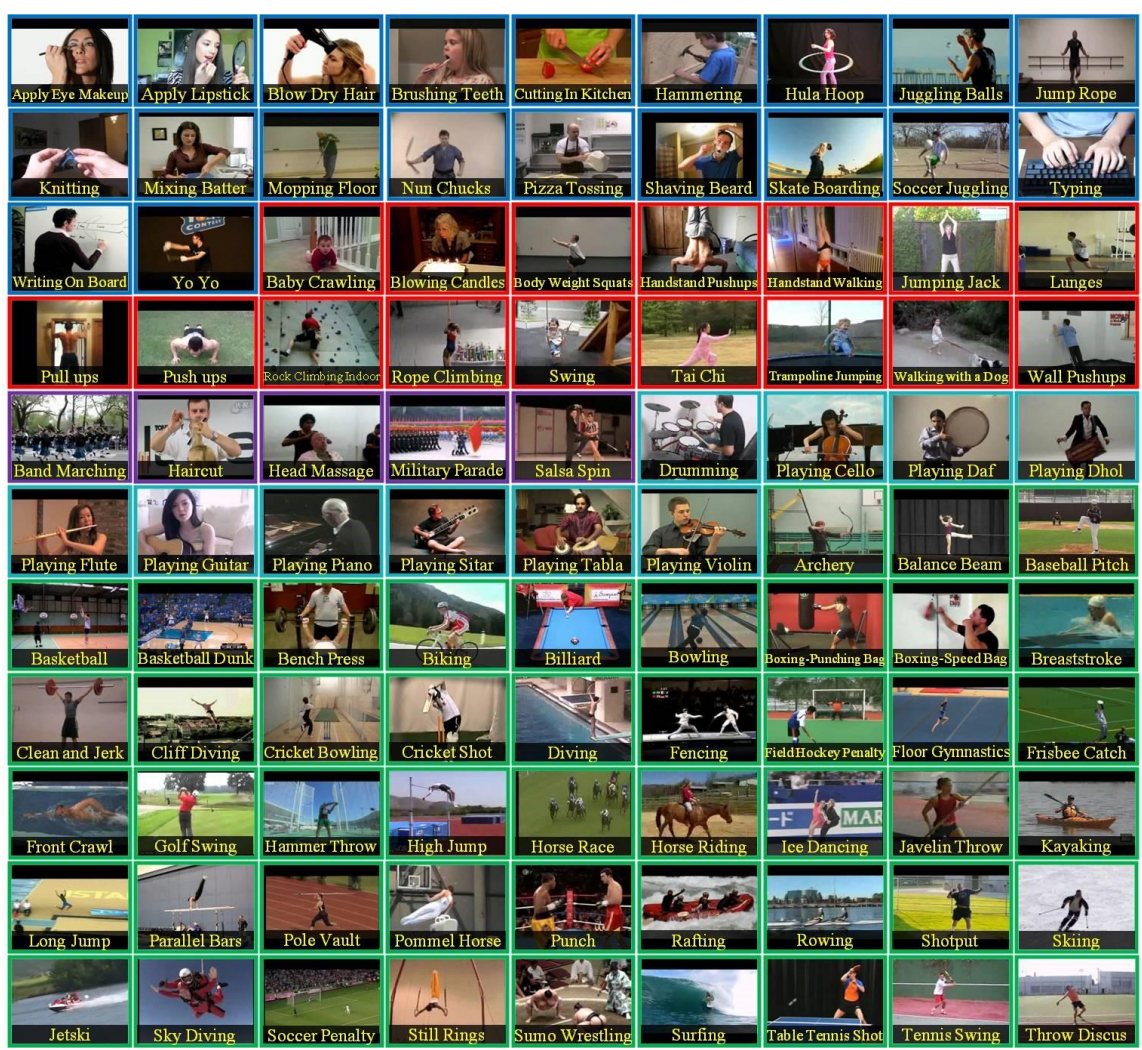

**Figure 4.** Dataset Sample for UCF 101 Classes Training and Testing Split.

Recognition and recognition are the base for providing categories for video data classification. The THUMOS challenge is one that provides solutions from large scale action recognition with a large number of classes. To keep the reported results consistent, we followed "Three Train/Test Splits" in experimental setup. The splitting eliminates randomness in the experimental setup. The train test split of the dataset followed the guidelines provided by the THUMOS challenge [1].

*3.2. Convolution Layer and Fully Connected Layer*

The convolution layer is used to convert representations through kernels into processable results. In any analysis, a kernel is used for a particular purpose. In a condensed phase, tiny areas of an image are inspected. The possibility of belonging to a filter class is given to a chart describing the enabled image layers. In the 3D CNN, kernels transfer the data (height, width, duration, and depth) in three dimensions and generate 3D maps for activation (Figure 5). The three-dimensional model for functional extraction is shown in Figure 5. Pooling Layer: The activation maps produced during convolution shall be used for sampling or sampling. A filter passes through an activation map, which measures the convolution phase in a specific segment at a time. The average scanned area, a weighted average on a central pixel, or the maximum value is taken from this filter and summed in the current chart.

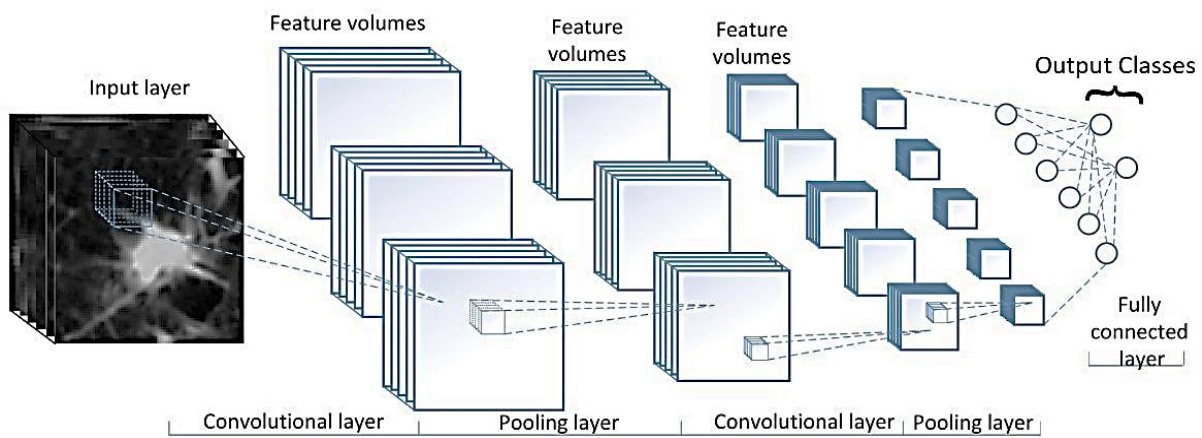

**Figure 5.** The three-dimensional model used for feature extraction.

The most frequently used max-pooling technique, in which the maximum value is obtained from the scanned region, works as a noise suppressor at compression. This abstraction decreases the computing power required for evaluating a map by extracting unimportant characteristics and making spatial variances possible, independent of rotation or tilting features. Ultimately linked (FC), a layer: The possibilities defined are evaluated, and the result is assigned a name, a logit, after many iterations, often thousands of convolutions and pooling of outgoing layers. This research is achieved by the Fully Connected layer, in which each flattened output layer handles interlocking nodes identical to a fully connected neural network (FCNN). The distinction is that the coagulation and pooling layers of a CNN are identical to the FC sheet. CNN will limit the need for greater processing power to the final steps by isolating the image features before feeding the output into the FC layer.

*3.3. Human Action Recognition*

Action recognition is the interpretation of object locations in a 2D picture set, such as a film, to either perceive or anticipate object behavior within the background of the scene frames. Action recognition and action detection are utilized in designing assistive technology such as intelligent residences, control software, or surveillance devices, and virtual reality applications, such as building shared conference rooms. This method is complicated by the need to consider arbitrary movements, such as camera or context objects, the speed of 3D conversion, and the lack of suitable parameter-modeling datasets. It is the most effective two-stream approach for evaluating spatial-temporal data in convolution and pooling layers separately for merging them into the FC system. (Figure 6).

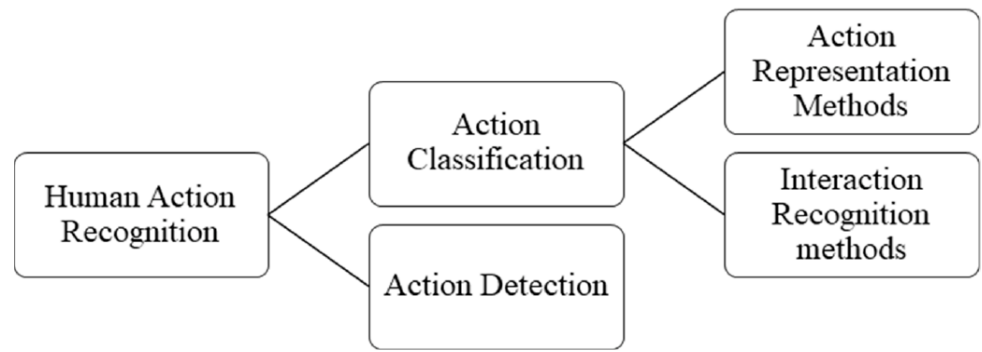

**Figure 6.** A classification framework for human action recognition methods.

If we look into the data types involved, the task of human action recognition can be accomplished by methods that can be classified into methods based on depth data and on color [40–44] (RGB). The framework for classifying such methods is given in Figure 3.

Following the advancement in machine learning, these methods can be further categorized into methods with hand-designed features which use traditional machine learning or methods with deep end-to-end learning for automatic feature extraction. The main aim is to extract human action features which in our case uses deep learning to extract STIP (Spatio-temporal interest point) features. This is because hand-crafted features limit the performance of human action recognition due to many factors in video, e.g., cinematography (movement of camera), occlusion, and pose estimation.

## 4. Experimental Details and Results

The specifics surrounding 3D CNN deployment with the experimental details of the UCF101 datasets have been carried out, as shown in Figure 4. We execute tests of three train/test breaks, the most common split also used in the THUMOS challenge [1,14]. The effects are calculated by the precision of the classification for each segment and by fine-tuning the state-of-the-art video classification structures for the suggested preprocessing methodology.

Experimental Equipment and setup: Our solution to a heterogeneous configuration of Processor-GPU has been tested, which involves the Intel Core i5-4460 Chip and the NVidia GeForce GTX 1080 Ti GPU. Table 2 displays the characteristics of the computer.

**Table 2.** Hardware used for the experimental setup.

| Sr. # | Device | CPU | GPU |
|---|---|---|---|
| 1. | Architecture | Ice Lake | Pascal |
| 2. | Base Clock | 3.2 GHz | 0.980 GHz |
| 3. | Boost Clock | 3.4 GHz | 1.033 GHz |
| 4. | Total Cores | 6 | 4400 (CUDA cores) |
| 5. | Memory | 16 GB | 12 GB |
| 6. | Memory bandwidth | 25.6 GB/s | 192.2 GB/s |
| 7. | Performance (Single Precision) | 409.6 GFLOPS | 2257.9 GFLOPS |
| 8. | OpenCL SDK | Intel SDK for OpenCL 2016 | CUDA 8.0 |
| 9. | Compiler | GCC 5.4.0 | NVCC |

*4.1. Performance Metrics*

Several tests are used to test the efficiency of the proposed model. Precision, recall, and F1-Score are used to assess the classification success (represent the classification outcomes with a focus on Recall) to calculate the success of the video classification. The dimensions are measured with the equations below. Figure 7 illustrates the theoretical model's confusion matrix.

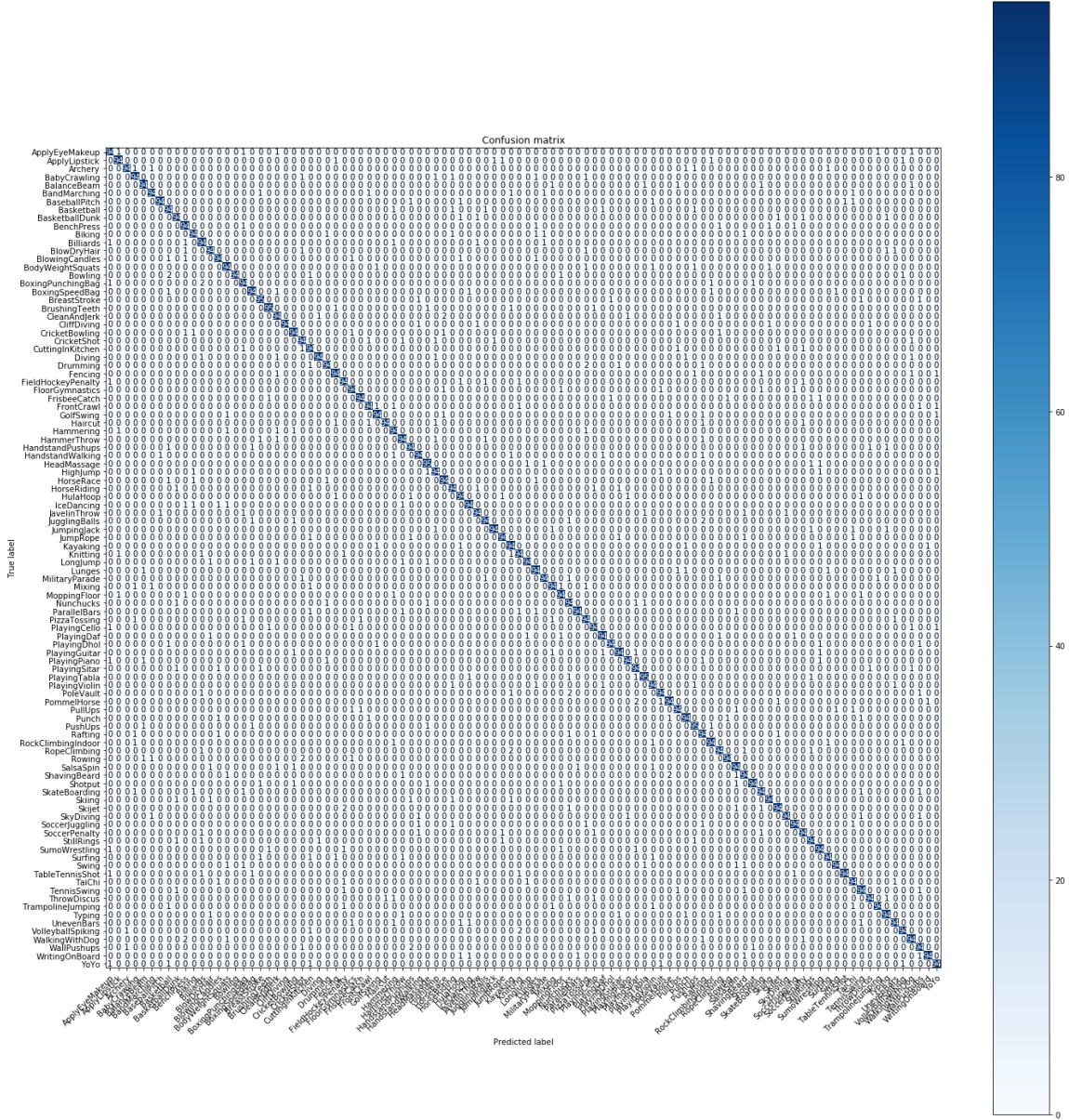

**Figure 7.** Confusion Matrix.

*4.2. Accuracy*

Precision applies to the proportion of the correctly graded sample, as seen in (1).

$$\text{Accuracy} = \frac{\text{No. of Correctly classified sample}}{\text{Total Sample}} \quad (1)$$

Figure 8 represents the accuracy values of ten groups that compare the proposed technique with Accuracy, Recall, and F-1 Score. We calculate the accuracy values for all the class groups through equation one. The F-1 Score through circus graphical representation shows accuracy values for every single group class of video classification.

The precision values presented in the figure show the samples classified through the provided dataset. Here, accuracy, precision, and Recall are the three main models used to compare the system accuracy.

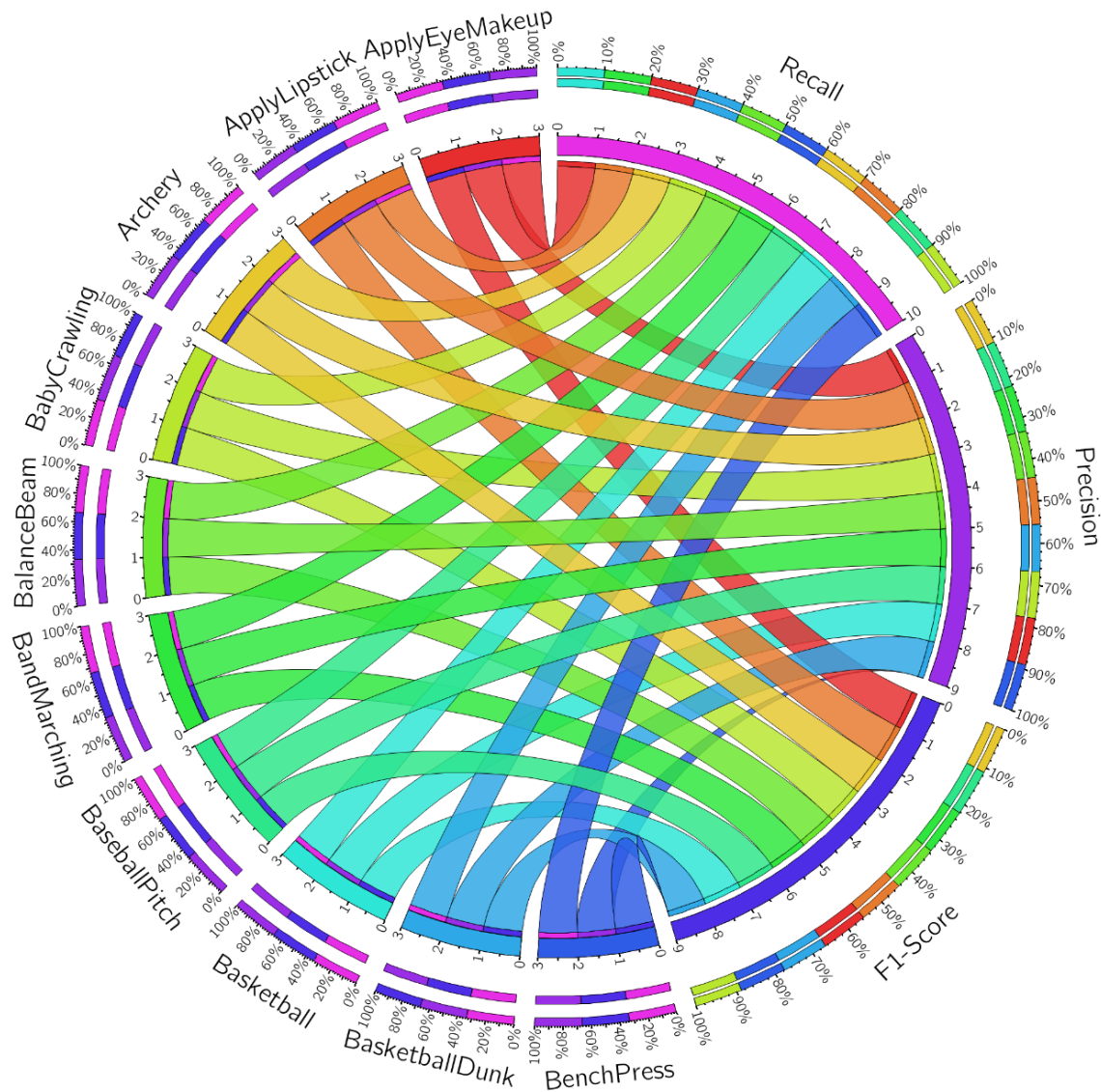

**Figure 8.** Results represent the accuracy values of ten groups.

*4.3. Precision*

Precision is a perfect test to assess whether false-positive losses are high (2), e.g., identification of e-mail spam. A false positive indicates that a non-spam e-mail (real negative) has been marked as spam (predicted spam) in the e-mail spam identification. If the accuracy of the spam detection model is not good, the e-mail user can miss critical e-mails.

$$\text{Precision} = \frac{\text{True Positive}}{\text{True Positive} + \text{False Positive}}$$

(2)

*4.4. Recall*

Recall actually calculates how many of the Actual Positives our model predicts through labeling it as Positive (True Positive). Applying this understanding, we know that Recall should be the model metric we use to select the best model when there is a high cost associated with False Negatives. So, Recall here is of critical importance when compared with the models that do the same kind of work (human action recognition and classification).

$$\text{Recall} = \frac{\text{True Positive}}{\text{True Positive } + \text{ False Negative}} \quad (3)$$

*4.5. F1-Score*

The F1 Score is required for a compromise between accuracy and warning. We have shown before that consistency will also be added with a good amount of true negatives, which, in certain market conditions, we may not rely on a lot, while the falsified negatives and the false positives generally have costs for companies (tangible and intangible) as mentioned in (4).

$$\text{F1} = 2 * \frac{\text{Precision} * \text{Recall}}{\text{Precision} + \text{Recall}} \quad (4)$$

Some other classes are required to check the results of the proposed model with precision, Recall, and other F1-Score. According to the model presented in Figure 8, we successfully show that the proposed techniques' results are effectively better and improved upon the Precision, Recall, and F-1 Score levels. The results are classified for another ten video classes in Figure 9.

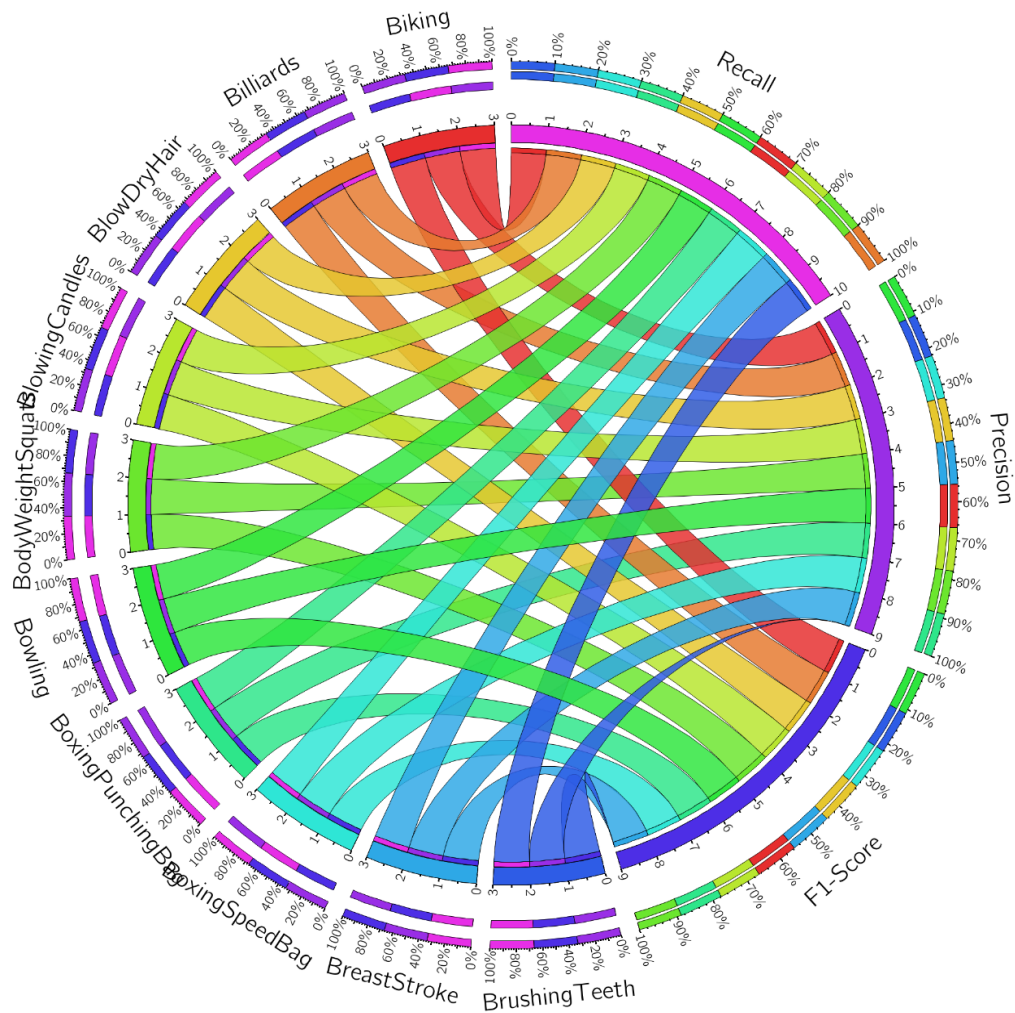

**Figure 9.** Results of the proposed technique against Recall, Precision, and F1-Score.

The accuracy is calculated from the confusion matrix, as shown in Figure 7. A comparison with the existing techniques is provided to evaluate the proposed method. Further, the proposed model performance parameters such as Support, Precision, Recall, and F1-Score are given in Table 3. The minimum value attained by the precision is 0.91, and the maximum value is 1.00, with an average of 0.9505, as shown in Table 3.

**Table 3.** Proposed model results for Precision, Recall, F1-Score and Support.

| Metric<br>Class | Precision | Recall | F1-Score | Support |
|---|---|---|---|---|
| ApplyEyeMakeup | 0.92 | 0.95 | 0.94 | 99 |
| ApplyLipstick | 0.96 | 0.95 | 0.95 | 99 |
| Archery | 0.98 | 0.95 | 0.96 | 99 |
| BabyCrawling | 0.94 | 0.95 | 0.94 | 99 |
| BalanceBeam | 0.96 | 0.95 | 0.95 | 99 |
| BandMarching | 0.96 | 0.95 | 0.95 | 99 |
| BaseballPitch | 0.98 | 0.95 | 0.96 | 99 |
| Basketball | 0.91 | 0.95 | 0.93 | 99 |
| BasketballDunk | 0.94 | 0.95 | 0.94 | 99 |
| BenchPress | 0.92 | 0.95 | 0.94 | 99 |
| Biking | 0.95 | 0.95 | 0.95 | 99 |
| Billiards | 0.91 | 0.95 | 0.93 | 99 |
| BlowDryHair | 0.95 | 0.95 | 0.95 | 99 |
| BlowingCandles | 0.95 | 0.95 | 0.95 | 99 |
| BodyWeightSquats | 0.94 | 0.95 | 0.94 | 99 |
| Bowling | 1 | 0.95 | 0.97 | 99 |
| BoxingPunchingBag | 0.92 | 0.95 | 0.94 | 99 |
| BoxingSpeedBag | 0.95 | 0.95 | 0.95 | 99 |
| BreastStroke | 0.95 | 0.96 | 0.95 | 99 |
| BrushingTeeth | 0.97 | 0.96 | 0.96 | 99 |
| CleanAndJerk | 0.92 | 0.95 | 0.94 | 99 |
| CliffDiving | 1 | 0.95 | 0.97 | 99 |
| CricketBowling | 0.95 | 0.95 | 0.95 | 99 |
| CricketShot | 0.93 | 0.95 | 0.94 | 99 |
| CuttingInKitchen | 0.91 | 0.95 | 0.93 | 99 |
| Diving | 0.98 | 0.95 | 0.96 | 99 |
| Drumming | 0.96 | 0.95 | 0.95 | 99 |
| Fencing | 0.95 | 0.95 | 0.95 | 99 |
| FieldHockeyPenalty | 0.93 | 0.95 | 0.94 | 99 |
| FloorGymnastics | 0.94 | 0.95 | 0.94 | 99 |
| FrisbeeCatch | 0.98 | 0.95 | 0.96 | 99 |
| FrontCrawl | 0.96 | 0.95 | 0.95 | 99 |
| GolfSwing | 0.96 | 0.95 | 0.95 | 99 |
| Haircut | 0.99 | 0.95 | 0.97 | 99 |
| Hammering | 0.92 | 0.95 | 0.94 | 99 |
| HammerThrow | 0.93 | 0.95 | 0.94 | 99 |
| HandstandPushups | 0.94 | 0.95 | 0.94 | 99 |
| HandstandWalking | 0.95 | 0.95 | 0.95 | 99 |
| HeadMassage | 0.94 | 0.96 | 0.95 | 99 |
| HighJump | 0.94 | 0.95 | 0.94 | 99 |
| HorseRace | 0.95 | 0.95 | 0.95 | 99 |
| HorseRiding | 0.94 | 0.95 | 0.94 | 99 |
| HulaHoop | 0.93 | 0.95 | 0.94 | 99 |
| IceDancing | 0.96 | 0.95 | 0.95 | 99 |
| JavelinThrow | 0.95 | 0.95 | 0.95 | 99 |
| JugglingBalls | 0.96 | 0.95 | 0.95 | 99 |
| JumpingJack | 0.98 | 0.95 | 0.96 | 99 |
| JumpRope | 0.96 | 0.95 | 0.95 | 99 |
| Kayaking | 0.92 | 0.95 | 0.94 | 99 |
| Knitting | 0.94 | 0.95 | 0.94 | 99 |
| LongJump | 0.97 | 0.95 | 0.96 | 99 |
| Lunges | 0.93 | 0.95 | 0.94 | 99 |
| MilitaryParade | 0.96 | 0.95 | 0.95 | 99 |
| Mixing | 0.95 | 0.95 | 0.95 | 99 |
| MoppingFloor | 0.94 | 0.95 | 0.94 | 99 |
| Nunchucks | 0.92 | 0.95 | 0.94 | 99 |
| ParallelBars | 0.98 | 0.95 | 0.96 | 99 |
| PizzaTossing | 0.92 | 0.95 | 0.94 | 99 |
| PlayingCello | 0.94 | 0.95 | 0.94 | 99 |
| PlayingDaf | 0.96 | 0.95 | 0.95 | 99 |
| PlayingDhol | 0.98 | 0.95 | 0.96 | 99 |
| PlayingGuitar | 0.96 | 0.95 | 0.95 | 99 |
| PlayingPiano | 0.97 | 0.95 | 0.96 | 99 |
| PlayingSitar | 0.94 | 0.95 | 0.94 | 99 |
| PlayingTabla | 0.96 | 0.96 | 0.96 | 99 |
| PlayingViolin | 0.94 | 0.95 | 0.94 | 99 |
| PoleVault | 0.96 | 0.95 | 0.95 | 99 |
| PommelHorse | 0.96 | 0.95 | 0.95 | 99 |
| PullUps | 0.95 | 0.95 | 0.95 | 99 |
| Punch | 0.95 | 0.95 | 0.95 | 99 |
| PushUps | 0.94 | 0.96 | 0.95 | 99 |
| Rafting | 0.9 | 0.95 | 0.93 | 99 |
| RockClimbingIndoor | 0.95 | 0.95 | 0.95 | 99 |
| RopeClimbing | 0.94 | 0.95 | 0.94 | 99 |
| Rowing | 0.98 | 0.95 | 0.96 | 99 |
| SalsaSpin | 0.94 | 0.95 | 0.94 | 99 |
| ShavingBeard | 0.93 | 0.95 | 0.94 | 99 |
| Shotput | 0.98 | 0.95 | 0.96 | 99 |
| SkateBoarding | 0.95 | 0.95 | 0.95 | 99 |
| Skiing | 0.97 | 0.95 | 0.96 | 99 |
| Skijet | 0.96 | 0.95 | 0.95 | 99 |
| SkyDiving | 0.98 | 0.95 | 0.96 | 99 |
| SoccerJuggling | 0.96 | 0.95 | 0.95 | 99 |
| SoccerPenalty | 0.94 | 0.95 | 0.94 | 99 |
| StillRings | 0.95 | 0.95 | 0.95 | 99 |
| SumoWrestling | 0.93 | 0.95 | 0.94 | 99 |
| Surfing | 0.96 | 0.95 | 0.95 | 99 |
| Swing | 0.98 | 0.95 | 0.96 | 99 |
| TableTennisShot | 0.99 | 0.95 | 0.97 | 99 |
| TaiChi | 0.93 | 0.95 | 0.94 | 99 |
| TennisSwing | 0.93 | 0.95 | 0.94 | 99 |
| ThrowDiscus | 0.97 | 0.95 | 0.96 | 99 |
| TrampolineJumping | 0.98 | 0.95 | 0.96 | 99 |
| Typing | 0.95 | 0.95 | 0.95 | 99 |
| UnevenBars | 0.95 | 0.95 | 0.95 | 99 |
| VolleyballSpiking | 0.96 | 0.95 | 0.95 | 99 |
| WalkingWithDog | 0.93 | 0.95 | 0.94 | 99 |
| WallPushups | 0.92 | 0.95 | 0.94 | 99 |
| WritingOnBoard | 0.98 | 0.95 | 0.96 | 99 |
| YoYo | 0.95 | 0.95 | 0.95 | 99 |
| **Average** | **0.9505** | **0.9505** | **0.9485** | **99** |

While Recall shows that the min value is 0.95, and the max value is 0.96, with an average of 0.9505, the F1-Score has a min value of 0.93, and the max value is o.97, with an average calculated of 0.9485. the overall support measured during the experimentation is 99. The support is the number of occurrences of each class in an actual set. It is clear from the comparison that the proposed technique is well performed and more accurate, as shown in Table 4.

**Table 4.** Results comparison with existing techniques.

| References | Dataset | Data Augmentation | Model Selection | Traditional ML Classifier | Deep Learning Framework | Output Classes | Classification Accuracy |
|---|---|---|---|---|---|---|---|
| Donahue et al. [34] | UCF 101 | No | Yes | SVM | No | 101 | 82.9% |
| Srivastava et al. [33] | UCF 101 | No | Yes | Genetic Fuzzy | No | 101 | 84.3% |
| Wang et al. [31] | UCF 101 | No | - | SVM | - | 101 | 85.9% |
| Tran et al. [35] | UCF 101 | Yes | Yes | Decision Tree | - | 101 | 86.7% |
| Simonyan et al. [36] | UCF 101 | No | Yes | Fuzzy | - | 101 | 88.0% |
| Lan et al. [19] | UCF 101 | No | Yes | Random Forest | No | 101 | 89.1% |
| Zha et al. [37] | UCF 101 | No | - | SVM | No | 101 | 89.6% |
| Zhou et al. [38] | UCF 101 | Yes | Yes | SVM | Yes | 101 | 94.0% |
| Proposed Method (HDLF) | UCF 101 | Yes | Yes | DNN | Yes | 101 | 95.0% |

Figure 10 describes the results comparison of the proposed technique with previous techniques, which illustrates that the proposed technique performs better than existing techniques with 95% accuracy. The minimum accuracy achieved during the implementation is 93%, and the maximum accuracy is 99%, with an average accuracy of 95%. The results table in Figure 9 compares the result of our experiment with previously used techniques visually.

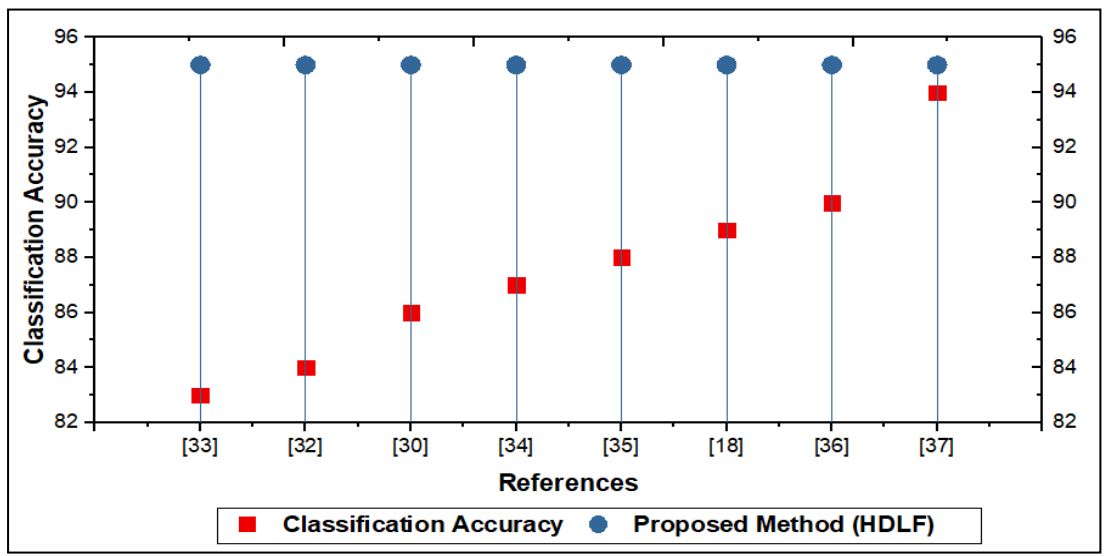

**Figure 10.** Results Comparison.

## 5. Threats and Limitations of Research

Temporal relations are of high importance in video data. So, in the future, it would be favorable to consider a deeper integration of information from a temporal sequence in the video into the 3D CNN itself. Undertaking this may yield better results when dealing with more extensive datasets such as the new kinetics 700 datasets. This research is limited to add only CNN-based video streaming data. The data used in this research are auto-captured from the video taken from different sources. According to limitations, this research does not handle the data generated from real-time IoT devices such as CCTV cameras, sensors, or audio devices. So, according to our comments, these limitations are limited to 3D multimedia data only. The proposed model can be used on bigger datasets such as kinetic 700, but it would require a lot of storage space for videos, and it will also require a lot of computational power to train the model on such a big dataset.

## 6. Conclusions

This paper investigates the comprehensive 3D CNN pipeline for segmenting actions/objects in videos and classifying the videos. This technique uses a pre-trained model for a broad measurement role as an encoder such that we undertake unattended video classification. We use different filters to substantially reduce the device load of the regular 3D convolutions. Extensive research on various recent techniques is being applied to generate classification models for the datasets. The power of our approach to Spatio-temporal intervention and video detection is revealed in contrast with state-of-the-art approaches. We proposed a competitively powerful, profound video classification model that integrates spatial and temporal features with the cited 3D ConvNet applications. It seems that 3D ConvNet training on STIP is far easier than training on raw stacked images. The latter is too complex, and architectural reform may be required. As we have seen, additional training data is helpful to our Spatio-Timed ConvNet, so continue to focus on broad video datasets such as the Kinetics-700 series. However, that provides an immense amount of training data in TBs, which creates a significant challenge. The new shallow image, absent in our present architecture, still has some essential ingredients in it. The most famous is a spatial phenomenon that is based on the trajectories via Spatio-temporal tubes. Since each entry captures the optical movement along the paths, our network does not provide spatial pooling of the trajectories. The specific treatment of camera motions is another possible enhancement field. It is of utmost importance to develop the capacity to model the temporal component of images in the future. Moreover, audio features proven to be helpful for scoring videos can easily be integrated into our system. In the future, improvements and the capability to model videos' temporal dimensions are highly prioritized. In addition, audio features that are useful for video classification can be easily incorporated into our framework.

**Author Contributions:**

**Funding:**

**Acknowledgments:**

**Conflicts of Interest:** The authors anonymously declared no conflicts of interest.